\pdfoutput=1

\documentclass[11pt]{article}

\usepackage{etex}
\reserveinserts{28}

\usepackage[preprint]{acl}

\usepackage{times}
\usepackage{latexsym}

\usepackage{booktabs}

\usepackage{pifont}

\usepackage{tcolorbox}

\usepackage{verbatim} 

\usepackage{geometry}
\usepackage{multirow}

\usepackage{amssymb}

\usepackage{enumitem}

\usepackage[T1]{fontenc}

\usepackage[utf8]{inputenc}

\usepackage{microtype}

\usepackage{inconsolata}

\usepackage{graphicx}  
\usepackage{caption}   
\usepackage{amsmath}   

\usepackage{subfigure}

\usepackage{authblk}

\newcommand{\xinhao}[1]{\textcolor{black}{#1}}

\title{Can We Edit LLMs for Long-Tail Biomedical Knowledge?}

\author[1]{Xinhao Yi}
\author[1]{Jake Lever}
\author[1]{Kevin Bryson}
\author[1]{Zaiqiao Meng}
\affil[1]{University of Glasgow}
\affil[ ]{\texttt{\{x.yi.2, jake.lever, kevin.bryson, zaiqiao.meng\}@glasgow.ac.uk}}

\begin{document}
\maketitle
\begin{abstract}
Knowledge editing has emerged as an effective approach for updating large language models (LLMs) by modifying their internal knowledge. 
However, their application to the biomedical domain faces unique challenges due to the long-tailed distribution of biomedical knowledge, where rare and infrequent information is prevalent.
%
In this paper, we conduct the first comprehensive study to investigate the effectiveness of knowledge editing methods for editing \textit{long-tail} biomedical knowledge. Our results indicate that, while existing editing methods can enhance LLMs' performance on \textit{long-tail} biomedical knowledge, 
their performance on long-tail knowledge remains inferior to that on high-frequency popular knowledge, even after editing.
Our further analysis reveals that long-tail biomedical knowledge contains a significant amount of one-to-many knowledge, 
where one subject and relation link to multiple objects.
This high prevalence of one-to-many knowledge limits the effectiveness of knowledge editing in improving LLMs' understanding of long-tail biomedical knowledge,  highlighting the need for tailored strategies to bridge this performance gap\footnote{Code: \url{https://github.com/xinhaoyi/edit_bio_long_tail}}.

\end{abstract}

\section{Introduction}

Recently, knowledge editing~\cite{meng2022locating,yao2023editing} has emerged as a promising approach to efficiently update large language models (LLMs) by injecting new knowledge into their internal knowledge~\cite{touvron2023llama, achiam2023gpt}. 
These methods have shown remarkable performance in enhancing LLMs' performance across {several} general-domain tasks, such as question answering (QA)~\cite{huang2023transformer}, knowledge injection~\cite{li2024pmet}, and knowledge reasoning~\cite{wang2023cross}. 

\begin{figure}[!t]%
    \centering
    \resizebox{1.0\linewidth}{!}{
    \includegraphics[width=0.5\textwidth]{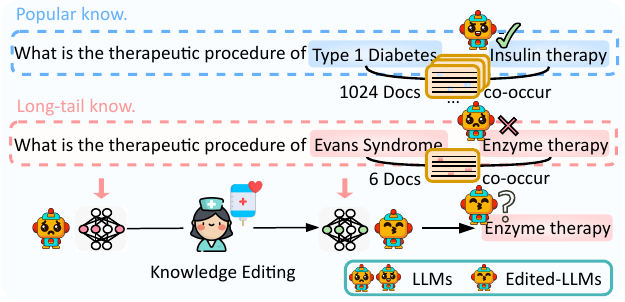}
    }
    \vspace{-1.5em}
    \caption{
    LLMs often struggle with long-tail biomedical knowledge, where entities co-occur in a few documents. 
    Knowledge editing offers a potential solution by injecting this rare information into LLMs, improving their ability to handle such long-tail knowledge. 
    }
    \label{fig:example_of_LLMs_struggle_with_long-tail_knowledge}
    \vspace{-1em}
\end{figure}

While knowledge editing methods have proven effective in general-domain tasks, their application to the biomedical domain presents unique challenges~\cite{wu2024medical}. 
Specifically, real-world biomedical data often exhibit a long-tailed distribution, with a small amount of popular knowledge and a large amount of long-tail knowledge that appears rarely or only once~\cite{wu2024medical,delile2024graph}. 
For example, the common disease ``Type 1 Diabetes'' is mentioned in over 106,138 papers in PubMed~\cite{roberts2001pubmed}, while a rare disease like ``Evans Syndrome'' appears in only about 23 papers~\cite{wei2013pubtator}. 
Recent studies indicate that the low frequency of knowledge in the pre-training corpus can hinder LLMs' understanding of this knowledge~\cite{kandpal2023large, wu2024medical}. 
Figure~\ref{fig:example_of_LLMs_struggle_with_long-tail_knowledge} illustrates an example where LLMs struggle with low-frequency biomedical knowledge. This is particularly problematic as LLMs are increasingly being used by healthcare professionals, including doctors, to assist in diagnosis and treatment recommendations~\cite{tian2024opportunities}. As LLMs become more integrated into clinical practice, their ability to accurately handle rare but critical biomedical knowledge becomes essential. 
This raises a critical question for knowledge editing in the biomedical domain: 

\textit{Can knowledge editing methods effectively edit large language models to incorporate long-tail biomedical knowledge?}

In this work, we present the first comprehensive 
study to investigate the effectiveness of knowledge editing for {long-tail} biomedical knowledge. 
We focus on biomedical knowledge represented as knowledge triples and leverage knowledge probing~\cite{alghanmi2021probing} to evaluate whether LLMs have effectively acquired this knowledge. Specifically, knowledge probing is a technique that queries LLMs to assess their internal factual knowledge~\cite{meng2021rewire}. 
As illustrated in Figure~\ref{fig:example_of_LLMs_struggle_with_long-tail_knowledge}, we probe LLMs with questions generated from biomedical knowledge triples to determine whether they can correctly recall the target knowledge. By comparing the knowledge probing results of LLMs before and after editing, we can evaluate how effectively knowledge editing enhances LLMs' ability to handle long-tail biomedical knowledge. Our key findings can be summarised as follows:

\begin{figure*}[!h]%
    \centering
    \resizebox{0.72\linewidth}{!}{
    \includegraphics[width=0.4\textwidth]{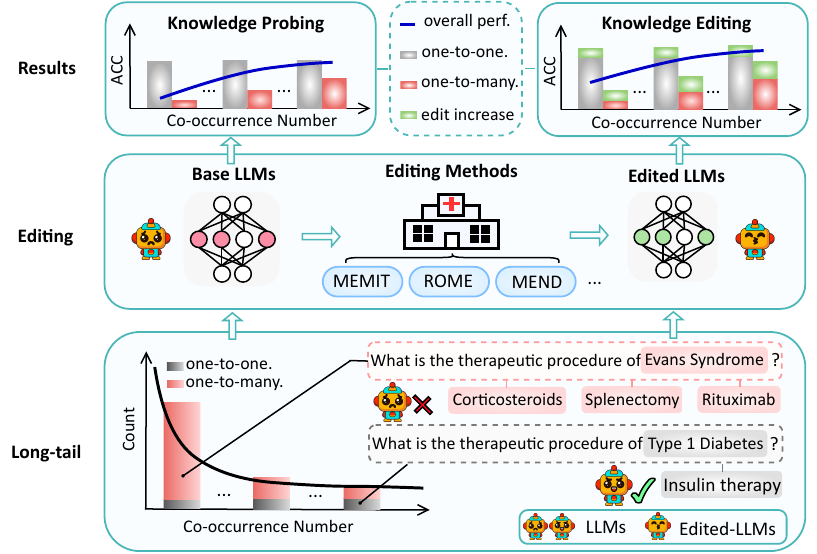}
    }
    \caption{
    An overview of probing and editing 
    for biomedical knowledge. These knowledge triples are classified into different groups based on co-occurrence number and further divided into one-to-one and one-to-many categories \xinhao{based on the number of correct answers} (see \S~\ref{sec:in-depth_analysis}). 
    \xinhao{The increasing performance with the number of co-occurrence number indicates that LLMs struggle to effectively capture long-tail biomedical knowledge before and after editing.} 
    }
    \label{fig:overall_structure}
\end{figure*}

{

\begin{itemize}
\vspace{-0.6em} \item LLMs struggle to capture long-tail biomedical knowledge through pre-training.
\vspace{-0.6em} \item Knowledge editing can improve LLMs’ performance on long-tail biomedical knowledge but remains less effective than on popular knowledge.
\vspace{-0.6em} \item Edited LLMs can memorise the form of long-tail knowledge, but their ability to generalise such knowledge is \xinhao{limited}. 
\vspace{-0.5em} \item The prevalence of one-to-many knowledge in long-tail biomedical knowledge is a key factor contributing to LLMs' poor performance in capturing 
\xinhao{such long-tail} knowledge.
\vspace{-0.6em} \item 
Effectively handling one-to-many knowledge is critical for improving LLMs' performance on long-tail biomedical knowledge through knowledge editing.
\end{itemize}

\section{Background and Definitions}

This section defines long-tail biomedical knowledge and briefly introduces the knowledge probing and editing techniques used in our experiments. 

\subsection{Long-Tail Biomedical Knowledge}
\label{sebsec:definition_long_tail}

We present biomedical knowledge using knowledge triple $\langle s, r, o \rangle$, where $s$ is the subject, $r$ is the relation, and $o$ is the object. 
Let $\mathcal{D}$ be the set of documents in the pre-training corpus, and $\mathcal{D}(s, o)$ be the subset of documents where both $s$ and $o$ co-occur. 
We define the \textit{co-occurrence number} of the knowledge triple as $|\mathcal{D}(s, o)|$, 
which represents the frequency of knowledge $\langle s, r, o \rangle$ within the document set $\mathcal{D}$~\cite{kandpal2023large}.
In this paper, following~\citet{mallen2022not} and~\citet{kandpal2023large}, we define \textit{long-tail knowledge} as:
\begin{equation}\label{eq:mf}
\mathcal{K}_{\text{l}} = \left\{ \langle s, r, o \rangle \mid |\mathcal{D}(s, o)| < \alpha \right\}, 
\end{equation}
where $\mathcal{K}_{\text{l}}$ denotes the set of long-tail knowledge and $\alpha$ represents a predefined threshold.

\subsection{Knowledge Probing}
\label{sebsec:knowledge_probing}

Knowledge probing aims to evaluate LLMs' ability to capture factual knowledge~\cite{meng2021rewire}, and can serve as \xinhao{an evaluation method to assess the effectiveness of knowledge editing~\cite{hernandez2023inspecting}.} 
Specifically, given a subject $s$ and a relation $r$ in a triple $\langle s, r, o \rangle$, 
we use a manually designed template $\mathcal{T}(s, r)$ to generate \xinhao{a natural language question}, which is then fed into an LLM $f_\theta$ to generate the object $o$ as the answer. 
Following the work of~\citet{meng2022locating} and~\citet{kassner2021multilingual}, accuracy (ACC) is used to evaluate the performance of LLM in recalling the correct target entity $o$, which is formulated as:
\begin{equation}\label{eq:mf}
\mathbb{E}_{\langle s, r, o \rangle \sim \mathcal{P}}  \mathbb{I}\left\{ \arg\max_{y} f_\theta(y \mid \mathcal{T}(s, r)) = o \right\},
\end{equation}
where \( \mathbb{E}_{\langle s, r, o \rangle \sim \mathcal{P}} \) denotes the expectation over a set of knowledge triples $\mathcal{P}$, 
\xinhao{$y$ indicates the predicted answer} and $\mathbb{I}\{\cdot\}$ is the indicator function. In this paper, we compare the knowledge probing results of LLMs before and after knowledge editing to investigate the effectiveness of editing methods in handling long-tail biomedical knowledge. 

\subsection{Knowledge Editing}

Knowledge editing~\cite{yao2023editing} aims to 
\xinhao{inject a} new knowledge $\langle s, r, o\rangle$ into an LLM through a specific edit descriptor $(x_e, y_e)$~\cite{yao2023editing}. 
Given a knowledge $\langle s, r, o \rangle$ for editing, $x_e$ can be formulated as $\langle s, r\rangle$, and $y_e = o$. The ultimate target of knowledge editing is to obtain an edited model $f_{\theta_e}$, which effectively integrates the intended modifications within the editing scope, while preserving the model's performance for out-of-scope unrelated facts: 
\vspace{-0.4em}

\begin{equation}\label{eq:mf}
f_{\theta_e}(x) = 
\begin{cases} 
y_e & \text{if } x \in I(x_e, y_e) \\
f_{\theta}(x) & \text{if } x \in O(x_e, y_e)
\end{cases}
\end{equation}

Here, the \textit{in-scope} set $I(x_e, y_e)$ includes $x_e$ and its equivalence neighborhood $N(x_e,y_e)$, which includes related input/output pairs. In contrast, the out-of-scope $O(x_e, y_e)$ contains inputs that are unrelated to the edit descriptor $(x_e, y_e)$.

\section{Identifying Long-Tail Biomedical Knowledge}
\label{sec:identify_biomedical_long-tail_knowledge}

Due to the lack of biomedical datasets specifically designed to evaluate long-tail knowledge, we develop a pipeline to extract such knowledge. In this section, we outline the procedures for extracting long-tail biomedical knowledge, with further details provided in Appendix~\ref{appendix:clikt}}. 
Specifically, we focus on biomedical knowledge represented as knowledge triples. 
We extract triples from SNOMED CT~\cite{donnelly2006snomed}, which is a large biomedical knowledge graph comprising over 
1.4 million
clinical triples~\cite{Benson2021}, and widely used for assessing LLMs' understanding of biomedical knowledge~\cite{meng2021rewire}. \xinhao{Following previous work~\cite{kandpal2023large}, we adopt the co-occurrence number—i.e., how often a triple's subject and object appear in the same document—as a proxy for knowledge popularity. To identify the long-tail knowledge within these triples, we use an entity linking pipeline to compute the co-occurrence number of each triple in the PubMed corpus\footnote{\url{https://pubmed.ncbi.nlm.nih.gov/}}, which is a widely used biomedical corpus for pre-training. 
In the entity linking pipeline, we first use PubTator~\cite{wei2013pubtator} to annotate entities in the PubMed corpus and then use SapBERT~\cite{liu2020self} to link knowledge triple entities to PubMed entities. }
Subsequently, we calculate the co-occurrence number for each triple. 
Long-tail knowledge is defined as triples with a co-occurrence number less than 10~\cite{kandpal2023large}. 

To evaluate LLMs’ ability to capture these triples, we generate question-answer pairs following~\citet{meng2022locating}. For each triple, we construct a question using the subject and relation, with the object serving as the answer. For example, for the triple $\langle$\textit{Diabetes, treated\_by, Insulin}$\rangle$, the corresponding QA pair is: \textit{What is Diabetes treated by? Answer: Insulin}. The statistics of our extracted data are presented in Table~\ref{tab:click_statistics} and the template for constructing questions is provided in Table~\ref{table:examples_of_relation_templates}. We refer to our dataset as CliKT (Clinical Knowledge Triples). Details of the construction process can be found in Appendix~\ref{appendix:clikt} and Figure~\ref{fig:clikt_construction}.

\section{Knowledge Editing for Long-Tail Biomedical Knowledge}
\label{section:LLMs_struggle_with_biomedical_long-tail_knowledge}

In this section, we investigate the effectiveness of knowledge editing methods in enhancing LLMs' ability to handle long-tail biomedical knowledge. Since some editing methods like MEND~\cite{mitchell2021fast} and IKE~\cite{zheng2023can} require additional training data, we follow~\citet{meng2022locating} to divide our CliKT dataset into training, validation, and test sets (See Table~\ref{tab:click_statistics}), and report the results on the test set. Specifically, we detail the experimental setup in \S~\ref{subsec:experiment_setup}, and introduce the results of LLMs before and after editing in \S~\ref{subsec:pre_edit_results} and \S~\ref{subsec:post_edit_results}, respectively.

\begin{table}[tb]
\centering
\resizebox{0.95\linewidth}{!}{%
\begin{tabular}{lccc}
\hline
\textbf{Item}                   & \textbf{Train} & \textbf{Valid} & \textbf{Test} \\ \hline
\textbf{\# Triples}                       & 59,705          & 14,087         & 28,375          \\ 
\quad ${|\mathcal{D}(s, o)| <10^1 }$ & 52,297          & 11,476         & 22,952          \\
\quad ${|\mathcal{D}(s, o)| \in [10^1, 10^2)}$ & \textcolor{white}{0}5,363 & \textcolor{white}{0}2,055 & \textcolor{white}{0}4,110          \\
\quad ${|\mathcal{D}(s, o)| \in [10^2, 10^3)}$ & \textcolor{white}{0}1,659 & \textcolor{white}{00,}551 & \textcolor{white}{0}1,103           \\
\quad ${|\mathcal{D}(s, o)| \geq 10^3}$ & \textcolor{white}{00,}386 & \textcolor{white}{00,}105 & \textcolor{white}{00,}210 \\ \midrule
\textbf{\# Relations} & \textcolor{white}{00,0}21 & \textcolor{white}{00,0}21 & \textcolor{white}{00,0}21             \\
\textbf{\# Subjects} & 39,654 & 12,267 & 21,872          \\
\textbf{\# Objects} & \textcolor{white}{0}7,867 & \textcolor{white}{0}3,526 & \textcolor{white}{0}4,706          \\ \hline
\end{tabular}%
}
\vspace{-0.1em}
\caption{The statistics of CliKT dataset. ${|\mathcal{D}(s, o)|}$ represents the oc-occurrence number of knowledge triple.}
\label{tab:click_statistics}
\end{table}

\subsection{Experimental Setup}
\label{subsec:experiment_setup}

\noindent \textbf{LLMs.} 
In our experiments, we employ two widely used biomedical LLMs primarily pre-trained on the PubMed corpus: \textbf{BioGPT-Large}~\cite{luo2022biogpt} and \textbf{BioMedLM}~\cite{bolton2024biomedlm}. 
Additionally, we include four general-domain LLMs: \textbf{Llama2}~\cite{touvron2023llama}, \textbf{Llama3}~\cite{grattafiori2024llama}, \textbf{GPT-J}~\cite{wang2021gpt} and \textbf{Qwen2.5}~\cite{yang2024qwen2} to evaluate whether our findings generalise to models that are not specifically trained on biomedical data. Details of these LLMs are provided in Appendix~\ref{subsec:app_details_of_llms}. 

\vspace{0.5em}\noindent \textbf{Knowledge Editing Methods}. 
For knowledge editing, we employ the following methods, which have demonstrated strong effectiveness in knowledge injection tasks~\cite{wang2023knowledge}:
\begin{itemize}
\vspace{-0.5em} \item \textbf{ROME}~\cite{meng2022locating}: ROME updates an MLP layer to encode new information by treating the MLP module as a key-value memory. It relies on causal mediation analysis to precisely identify the location for editing.

\vspace{-0.5em} \item \textbf{MEMIT}~\cite{meng2022mass}: it employs the localisation strategies from ROME and applies explicit parameter adjustments to inject new knowledge across multiple layers.

\vspace{-0.5em} \item \textbf{MEND}~\cite{mitchell2021fast}: \xinhao{MEND enables} efficient, targeted updates to LLMs by leveraging low-rank gradient transformations. It enables quick, localised modifications in model behaviour using only a single input-output example, while preventing overfitting.

\vspace{-0.5em} \item \textbf{IKE}~\cite{zheng2023can}: \xinhao{IKE} modifies factual knowledge in LLMs through in-context learning without updating parameters. It corrects specific knowledge using demonstration contexts, reducing over-editing and preserving previously stored knowledge.

\vspace{-0.5em} \item \textbf{FT}~\cite{yao2023editing}: \xinhao{FT} updates model parameters using gradient descent on a single MLP layer identified by ROME. 
We employ the FT implementation within the EasyEdit framework~\cite{wang2023easyedit}. 

\end{itemize}

\begin{figure}[tb]%
    \centering
    \resizebox{0.95\linewidth}{!}{
    \includegraphics[width=0.4\textwidth]{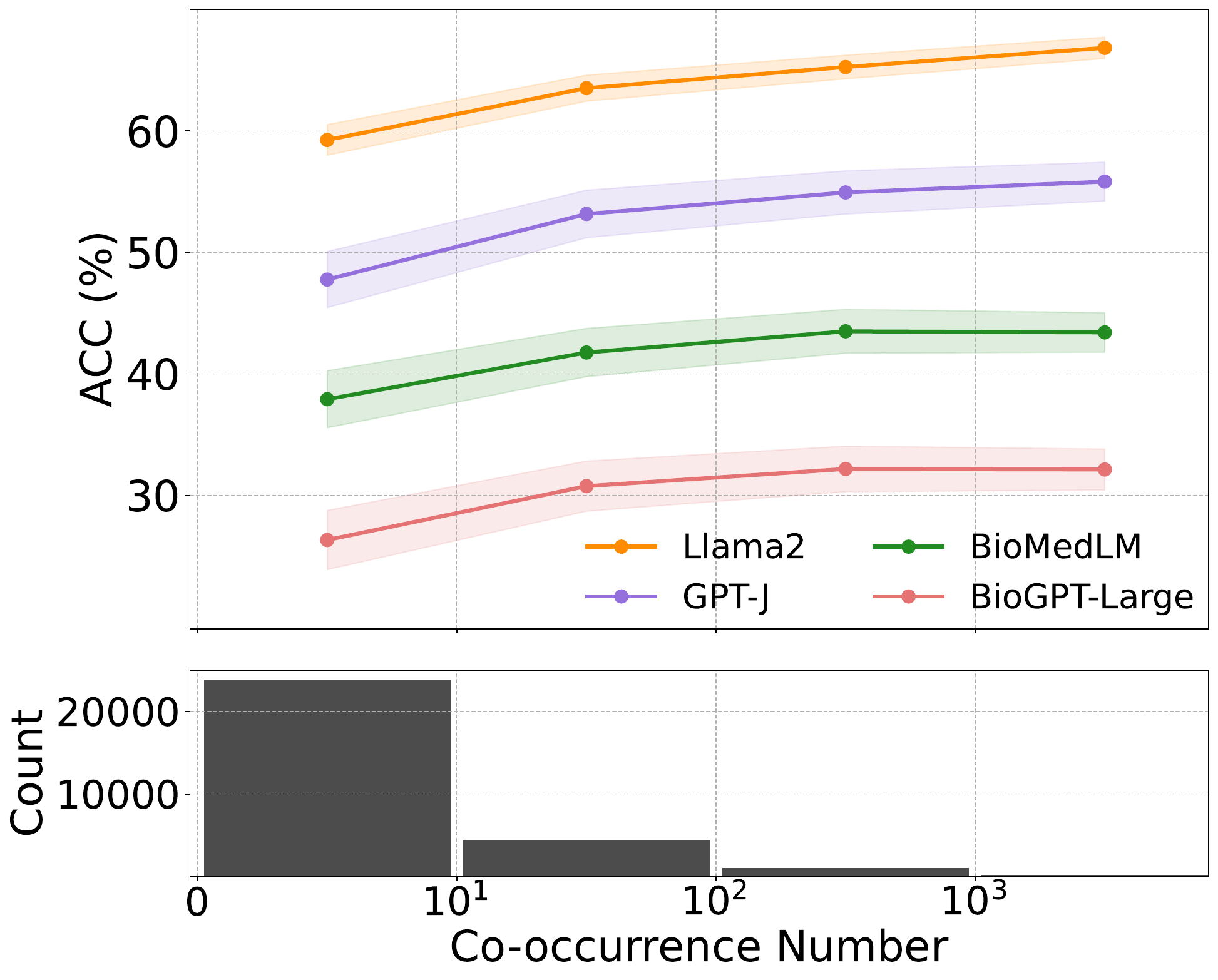}
    }
    \vspace{-0.5em}
    \caption{The overall performance of pre-edit probing on Llama2, GPT-J, BioMedLM and BioGPT-Large. The shaded areas indicate the standard deviation \xinhao{and Count denotes the number of triples within each group.}}
    \label{fig:probing_performance_on_long_tail_knowledge}
    \vspace{-0.5em}
\end{figure}

\noindent \textbf{Evaluation Metrics}.
We use knowledge probing to evaluate whether LLMs have successfully acquired biomedical knowledge within the CliKT dataset. Specifically, we focus on the zero-shot QA performance of LLMs in answering questions from the CliKT dataset. The questions are used as inputs, and the accuracy (ACC) metric is employed to evaluate the correctness of the generated answers, as described in \S~\ref{sebsec:knowledge_probing}.

In addition to knowledge probing, we follow previous works~\cite{meng2022locating,yao2023editing} and use the following metrics to evaluate the comprehensive effectiveness of knowledge editing: 
(1) \textbf{Reliability}: This metric measures the mean accuracy on a specific collection of pre-defined input-output pairs ($x_e$, $y_e$); 
(2) \textbf{Generalisation}: Considering that paraphrased sentences should be modified accordingly by editing, this metric measures the average accuracy on equivalent neighbours R($x_e$, $y_e$); 
(3) \textbf{Locality}: 
This metric quantifies how often the predictions of the post-edit model remain unchanged for out-of-scope neighbours O($x_e$, $y_e$). Detailed definitions of these metrics are provided in Appendix~\ref{subsec:details_of_evaluation_metrics}.

\begin{figure}[!t]%
    \centering
    \resizebox{1.0\linewidth}{!}{
    \includegraphics[width=0.4\textwidth]{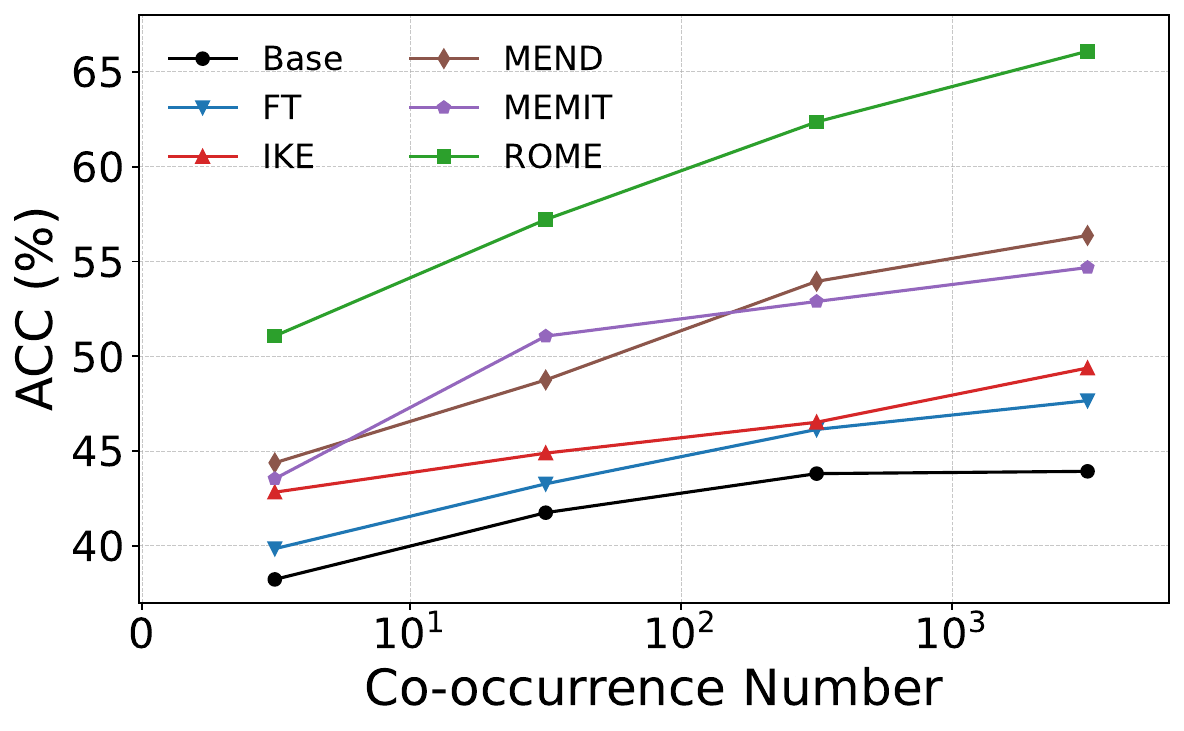}
    }
    \vspace{-1.5em}
    \caption{
    The performance of knowledge probing after editing with different editing methods on BioMedLM, where ``Base'' denotes LLM without editing. 
    }
    \label{fig:editing_performance}
    \vspace{-1.0em}
\end{figure}

\subsection{Pre-Edit Results on Long-Tail Biomedical Knowledge}
\label{subsec:pre_edit_results}

\vspace{0.5em} \noindent \textbf{Finding 1:} \textit{LLMs struggle to capture long-tail biomedical knowledge through pre-training.}
\vspace{0.5em}

\begin{figure*}[t]%
    \centering
    \resizebox{1.0\linewidth}{!}{
    \includegraphics[width=0.4\textwidth]{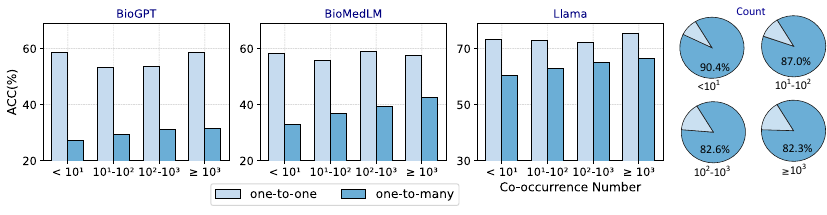}
    }
    \vspace{-1.5em}
    \caption{The comparison of knowledge probing performance between one-to-one and one-to-many settings across different co-occurrence numbers, with the pie chart on the far right illustrating the data distribution.}
    \label{fig:dif_relation_performance}
    \vspace{-0.5em}
\end{figure*}

To investigate whether LLMs face challenges in capturing long-tail biomedical knowledge during pre-training, we categorise biomedical knowledge triples  
in CliKT into different groups based on their co-occurrence number $|\mathcal{D}(s, o)|$ and evaluate the probing results of LLMs across these groups. 

The bottom portion of Figure~\ref{fig:probing_performance_on_long_tail_knowledge} shows the distribution of triples across the different groups, which 
\xinhao{highlights} the long-tail nature of biomedical knowledge, where long-tail knowledge accounts for the majority of the data. 
The results for biomedical LLMs and general-domain LLMs are illustrated in the top portion of Figure~\ref{fig:probing_performance_on_long_tail_knowledge}. 
Specifically, Figure~\ref{fig:probing_performance_on_long_tail_knowledge} shows that the performance of LLMs declines as the co-occurrence number decreases. 
In particular, the performance of BioMedLM on long-tail knowledge ($|\mathcal{D}(s, o)|<10$) is $22.86\%$ lower \xinhao{relative to} its performance on popular knowledge ($|\mathcal{D}(s, o)|\geq10^3$). 
This trend is also evident in general-domain LLMs. 
\xinhao{For example,} Llama2 experiences an accuracy drop of $16.86\%$ when handling long-tail biomedical knowledge compared with popular knowledge. 
These results indicate that 
LLMs struggle with long-tail biomedical knowledge, highlighting the challenge of accurately capturing long-tail knowledge during pre-training. 
%
\xinhao{Furthermore, Figure~\ref{fig:probing_performance_on_long_tail_knowledge} shows that as the co-occurrence number decreases, the standard deviation of ACC increases.} 
This observation implies that LLMs exhibit greater confidence when processing popular biomedical knowledge \xinhao{than long-tail biomedical knowledge. }

Based on the above analysis, we conclude that LLMs indeed struggle to capture long-tail biomedical knowledge. 
As long-tail knowledge constitutes the majority of biomedical data, it is crucial to explore methods that can effectively improve LLMs' performance on long-tail biomedical knowledge. 

\begin{table}[t!]
\centering
\resizebox{0.85\linewidth}{!}{%
\begin{tabular}{llccc}
\toprule
\textbf{Group} & \textbf{Edit} & \textbf{Reliability$\uparrow$} & \textbf{Gen.$\uparrow$} & \textbf{Locality$\uparrow$} \\ \midrule
\multirow{5}{*}{<10$^1$}       & ROME   & \textbf{98.02} & \textbf{68.42} & 83.70 \\
                               & MEMIT  & 86.21 & \underline{47.36} & \textbf{98.10} \\
                               & MEND    & \underline{91.32} & 46.75 & 89.60 \\
                               & IKE   & 83.87 & 43.70 & \underline{97.81} \\
                               & FT     & 32.52 & 40.36 & 96.80 \\ \midrule
\multirow{5}{*}{[10$^1$, 10$^2$)} & ROME   & \textbf{98.11} & \textbf{70.10} & 84.60 \\
                               & MEMIT  & \underline{89.21} & \underline{48.21} & \underline{97.30} \\
                               & MEND    & 88.90 & 47.80 & 89.83 \\
                               & IKE   & 84.52 & 45.12 & 96.80 \\
                               & FT     & 33.35 & 40.78 & \textbf{97.90} \\ \midrule
\multirow{5}{*}{[10$^2$, 10$^3$)} & ROME   & \textbf{98.63} & \textbf{72.50} & 84.62 \\
                               & MEMIT  & \underline{89.01} & \underline{51.47} & \textbf{97.90} \\
                               & MEND    & 88.94 & 48.83 & 91.40 \\
                               & IKE   & 85.89 & 46.74 & \underline{96.85} \\
                               & FT     & 33.89 & 44.62 & 96.66 \\ \midrule
\multirow{5}{*}{$\geq$ 10$^3$} & ROME   & \textbf{98.66} & \textbf{72.54} & 84.45 \\
                               & MEMIT  & 89.87 & \underline{50.00} & \underline{97.43} \\
                               & MEND    & \underline{90.96} & 49.86 & 90.92 \\
                               & IKE   & 85.91 & 48.76 & 96.87 \\
                               & FT     & 34.84 & 44.62 & \textbf{97.57} \\ \bottomrule
\end{tabular}%
}
\vspace{-0.5em}
\caption{Performance of knowledge editing methods on the CliKT dataset across different co-occurrence number groups. The best performance per group is marked in boldface, while the second-best performance is underlined. 
$\uparrow$ indicates that higher values reflect better performance, and ``Gen.'' stands for Generalisation.}
\label{tab:other_editing_performance}
\end{table}

\subsection{Post-Edit Results for Long-Tail Biomedical Knowledge}
\label{subsec:post_edit_results}

\vspace{0.5em} 
\noindent \textbf{Finding 2:} \textit{Knowledge editing can improve LLMs’ performance on long-tail biomedical knowledge but remains less effective than on popular knowledge.}
\vspace{0.5em}

Subsequently, we investigate the effectiveness of knowledge editing for long-tail biomedical knowledge. We apply existing knowledge editing methods to inject biomedical knowledge from the CliKT dataset into LLMs and then follow the procedures in the pre-edit experiments for evaluation. 

The post-edit probing results for BioMedLM are presented in Figure~\ref{fig:editing_performance}, while the results for other LLMs can be found in Figure~\ref{fig:additional_editing_performance}. These results yield the following findings: 
(1) Knowledge editing methods, especially ROME, can enhance LLM's ability in handling long-tail biomedical knowledge. For example, Figure~\ref{fig:editing_performance} shows that BioMedLM edited with ROME achieves an improvement of approximately 52.08\% in ACC on long-tail knowledge ($|\mathcal{D}(s, o)|<10$) compared to the base model before editing;
(2) Despite the improvements from knowledge editing, Figure~\ref{fig:editing_performance} also reveals that ACC of post-edit LLMs consistently drops as the co-occurrence number decreases across all the editing methods.
Specifically, for ROME, the ACC on long-tail knowledge is still 16.15\% \xinhao{relatively} lower than on popular knowledge ($|\mathcal{D}(s, o)|\geq10^3$). 
This indicates that even after editing, the edited LLMs continue to suffer from long-tail biomedical knowledge.

\vspace{0.5em} 
\noindent \textbf{Finding 3:} \textit{Edited LLMs can memorise the form of long-tail knowledge, but their ability to generalise such knowledge is \xinhao{limited}.}
\vspace{0.5em}

In addition to the post-edit probing results, we also calculate the other editing metrics outlined in \S\ref{subsec:experiment_setup} to comprehensively evaluate the effectiveness of the editing methods. 
Specifically, we calculate the Reliability, Generalisation and Locality metrics of edited models across different groups of biomedical knowledge. From the results in Table~\ref{tab:other_editing_performance}, we observe that ROME’s Reliability remains above 98\% across all groups, with no significant variation. Similarly, the Reliability of MEMIT, MEND, and IKE is largely unaffected by the co-occurrence number, indicating that the edited LLMs’ ability to memorise the form of inserted knowledge is not influenced by long-tail knowledge. 
However, the generalisation performance declines as the co-occurrence number decreases, which aligns with the observed reduction in post-edit ACC for edited-LLMs as the co-occurrence number decreases.
This observation suggests that, although edited LLMs can memorize the form of long-tail knowledge itself after knowledge editing, 
their ability to generalise this long-tail knowledge, especially in reasoning and responding to related questions, remains influenced by low co-occurrence numbers. 

Furthermore, we observe that, though all the editing methods exhibit relatively strong performance in terms of locality across groups, ROME is affected more than the other methods. This indicates that while ROME achieves the best reliability and generalisation, it may slightly affect unrelated knowledge, consistent with the observations of Wang et al.~\cite{wang2024wise}.

\begin{figure*}[!h]%
    \centering
    \resizebox{0.7\linewidth}{!}{
    \includegraphics[width=0.4\textwidth]{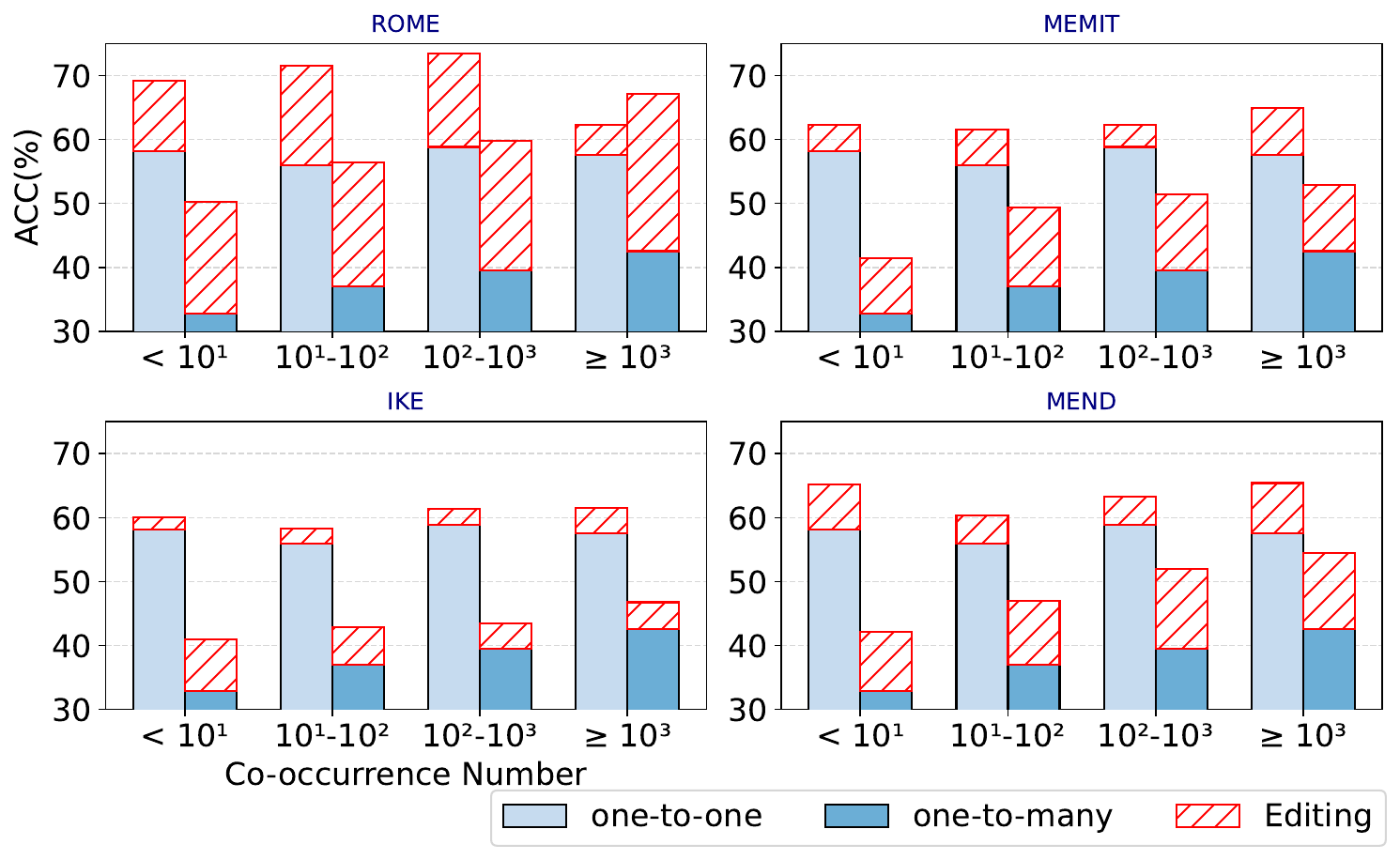}
    }
    \vspace{-0.5em}
    \caption{The knowledge probing performance of BioMedLM on both one-to-one knowledge and one-to-many knowledge before and after editing. }
    \vspace{-0.7em}
    \label{fig:dif_relation_editing_performance}
\end{figure*}

\subsection{In-depth Analysis of Knowledge Type in Knowledge Editing}
\label{sec:in-depth_analysis}
In this section, to further investigate the cause of the performance gap between long-tail and popular biomedical knowledge before and after editing, we 
further subdivide the data of long-tail and popular knowledge into \textit{one-to-one} and \textit{one-to-many} knowledge \xinhao{categories}. The \textit{one-to-one} knowledge means the subject is linked to a single object through the same relation, and \textit{one-to-many} knowledge means the subject is linked to multiple objects through the same relation. For example, 
the triple $\langle$\textit{Type 1 diabetes, therapeutic procedure, insulin therapy}$\rangle$ represents a one-to-one knowledge, where ``Type 1 diabetes'' is associated with a single object, ``insulin therapy''. In contrast, $\langle$\textit{hypertension, associated with, heart disease}$\rangle$ exemplifies a one-to-many knowledge, where ``hypertension'' can be linked to multiple objects, such as ``stroke'' or ``kidney disease''. 

\vspace{0.5em} 

\subsubsection{Pre-Edit Probing of Different Types of Knowledge}

\noindent \textbf{Finding 4:} \textit{The prevalence of one-to-many knowledge in long-tail biomedical knowledge is a key factor contributing to LLMs' poor performance in capturing \xinhao{such long-tail} knowledge}.

Figure~\ref{fig:dif_relation_performance} presents the pre-edit probing results of one-to-one and one-to-many knowledge across different co-occurrence number groups. 
We found that one-to-one knowledge is almost unaffected by co-occurrence numbers and consistently outperforms one-to-many knowledge in all groups. 
For instance, BioGPT achieves an ACC that is approximately 115.56\% higher on one-to-one knowledge compared to one-to-many knowledge. 
In contrast, for one-to-many knowledge, results from BioGPT, BioMedLM, and Llama2 all show a steady increase in ACC as the co-occurrence number increases. 
This suggests that co-occurrence number, or knowledge frequency, has a significant impact on LLMs' ability to accurately comprehend one-to-many knowledge. 
We further analysed the distribution of one-to-one and one-to-many knowledge. 
Figure~\ref{fig:dif_relation_performance} shows that as the co-occurrence number increases, the proportion of one-to-many knowledge decreases while one-to-one knowledge increases. In the long-tail knowledge \xinhao{group} ($|\mathcal{D}(s, o)|<10$), 90.4\% of the knowledge is one-to-many. This analysis reveals that LLMs' difficulty with long-tail biomedical knowledge before editing is primarily due to the large proportion of one-to-many knowledge, which is challenging for LLMs to comprehend, as it increases the probability that the correct answers will not align with the model's output.

\vspace{0.5em} 
\subsubsection{Knowledge Editing for Different Types of Knowledge}

\noindent \textbf{Finding 5:} \textit{Effectively handling one-to-many knowledge is critical for improving LLMs' performance on long-tail biomedical knowledge through knowledge editing}.

Next, we apply editing methods to both one-to-one and one-to-many knowledge. The results for BioMedLM are provided in Figure~\ref{fig:dif_relation_editing_performance}, while the results for other LLMs can be found in Figure~\ref{fig:biogpt_dif_relation_editing_performance}. 
As shown in Figure~\ref{fig:dif_relation_editing_performance}, while editing methods enhance performance on one-to-many knowledge, the improvement remains limited. For instance, in the ROME-edited BioMedLM for the long-tail knowledge ($|\mathcal{D}(s, o)| < 10$), the ACC for one-to-one knowledge was initially 42.19\% higher than that for one-to-many knowledge. After applying the editing, this gap decreased to 16.43\%. 
However, the persistent gap also highlights that even after editing, the model's performance on one-to-many knowledge, which constitutes the majority of long-tail knowledge, remains constrained. 
This finding suggests that \textit{despite knowledge editing can enhance LLMs' capability in handling one-to-many knowledge, there remains a challenge in bridging the performance gap between one-to-one and one-to-many knowledge}. This limitation is critical given that one-to-many knowledge constitutes the majority of long-tail knowledge. 

\section{Related Work}

\subsection{LLMs for the Biomedical Domain}

LLMs have made significant success in the biomedical domain, with an increasing variety of models contributing to advancements across different tasks~\cite{tian2024opportunities}. 
In the initial stages of their application, BERT~\cite{vaswani2017attention} and its variants, such as BioBERT~\cite{lee2020biobert} and ClinicalBERT~\cite{huang2019clinicalbert}, demonstrated notable improvements in named entity recognition and relation extraction when applied to large datasets such as PubMed and clinical notes~\cite{perera2020named, sun2021biomedical}.
GPT-based models, including GPT-J~\cite{wang2021gpt}, BioGPT~\cite{luo2022biogpt} and BioMedLM~\cite{bolton2024biomedlm}, further enhanced biomedical text generation and question answering~\cite{tian2024opportunities}. Recent LLMs like Llama~\cite{touvron2023llama}, Falcon~\cite{almazrouei2023falcon}, and Palm~\cite{chowdhery2023palm} have scaled transformer architectures to address more complex tasks, such as biomedical knowledge reasoning~\cite{wu2024pmc, watanabe2024empowerw} and assisting in clinical decision-making~\cite{sandmann2024systematic}. 
This work explores LLMs' performance on long-tail biomedical knowledge. We present the first study to investigate how long-tail knowledge impacts LLMs in knowledge editing, offering new insights into improving LLMs' handling of rare biomedical information through knowledge editing techniques.

\subsection{Knowledge Editing}
Knowledge editing methods can be broadly classified into three distinct categories~\cite{yao2023editing}: memory-based~\cite{zheng-etal-2023-edit}, meta learning~\cite{mitchell2021fast}, and locate-then-edit~\cite{meng2022locating}. 
Memory-based methods, like IKE~\cite{zheng-etal-2023-edit}, enhance LLMs with external memory modules to update knowledge without changing the model's parameters. 
Meta-learning approaches, such as KE~\cite{de2021editing}, train a hyper-network to generate updated weights. MEND~\cite{mitchell2021fast} improves on this by using low-rank gradient updates for more efficient model edits. However, meta-learning methods still require substantial computational resources and may unintentionally affect unrelated knowledge.

Locate-then-edit approaches aim for more targeted knowledge editing. Methods like KN~\cite{dai2021knowledge} use knowledge attribution to locate relevant neurons but struggle with precise weight updates. ROME~\cite{meng2022locating} advances this by using causal tracing to locate and edit the Feed Forward Network (FFN) layers, which act as key-value memories~\cite{geva2020transformer,geva2023dissecting}. MEMIT~\cite{meng2022mass} further expands this technique for batch editing. 
To the best of our knowledge, this work is the first to investigate the effectiveness of knowledge editing on long-tail biomedical knowledge.

\subsection{Long-Tail Knowledge within LLMs}

Existing studies have explored how long-tail knowledge, affects LLMs' performance~\cite{shin2022effect, han2022orca, elazar2022measuring, mallen2022not, kandpal2023large}.
\citet{mallen2022not} find that commonsense QA accuracy is strongly correlated with the frequency of entity popularity in the pre-training data from Wikipedia~\cite{milne2008learning}. Similarly, \citet{elazar2022measuring} employ causal inference to investigate how pre-training data statistics affect commonsense QA, highlighting how models rely on co-occurrence patterns between subjects, objects, and text to answer questions. More recently, \citet{kandpal2023large} explore the connection between the knowledge LLMs acquire for general-domain QA tasks and its frequency in the pre-training corpus, introducing comparative experiments involving model retraining and scaling. 

Despite these findings, prior work has focused on general-domain QA, with the long-tail biomedical domain remaining largely unexplored~\cite{wu2024medical}. This is especially concerning as LLMs are increasingly being used by healthcare professionals. Our research fills this gap by investigating the influence of long-tail biomedical knowledge on LLMs through knowledge probing and examining its impact on the effectiveness of knowledge editing.
This is particularly problematic as LLMs are increasingly being used by healthcare professionals, including doctors, to assist in diagnosis and treatment recommendations.

\section{Conclusion}

In this paper, we investigated the effectiveness of knowledge editing methods for addressing the challenges of long-tail biomedical knowledge in LLMs. Our findings show that while existing techniques enhance performance on long-tail knowledge, their performance remains inferior to that on high-frequency popular knowledge. This problem is primarily attributed to the high presence of one-to-many knowledge in the biomedical domain, which complicates the models’ ability to effectively comprehend such knowledge. To address these challenges, we recommend the development of advanced editing techniques specifically tailored to long-tail knowledge. These techniques should prioritise strategies for effectively handling the intricacies of one-to-many knowledge scenarios, which are particularly common in the biomedical domain and remain a significant obstacle for current methods.

\section*{Limitations}

We identify the following limitations of our work: (1) First, our approach to extracting long-tail knowledge is based on document-level co-occurrence frequency~\cite{kandpal2023large}, which captures general patterns of occurrence but lacks refinement at the sentence level. This limitation may cause our analysis to miss finer patterns in knowledge distribution, especially in instances where sentence-level context provides essential nuances. 
Future work could enhance the long-tail knowledge extraction pipeline by investigating co-occurrence on the sentence-level to improve the granularity of knowledge editing. (2)Second, our experimental framework is limited to the collection of over 100,000 biomedical knowledge extracted from PubMed, an extensive repository of biomedical literature. While we believe the scale of this collection offers a robust foundation for evaluating our methods, our future research should focus on extracting long-tail knowledge from a broader range of domains to further validate the generalisability of our findings. (3) Finally, we concentrate on analysing limitations without proposing specific solutions, prioritising the establishment of a comprehensive understanding. Future work will focus on developing methods to improve knowledge editing performance on long-tail knowledge.

\bibliography{custom}

\appendix

\newpage
\onecolumn  
\section*{Appendix}
\label{sec:appendix}

In the Appendix, we introduce more details along with dataset construction, additional experimental results, discussions, and related works:

\begin{itemize}
    \item \textbf{Appendix \ref{appendix:clikt}}: CliKT Construction (cf. Section 3).
    \item \textbf{Appendix \ref{appendix:experimental_details}}: Experimental Details (cf. Section 2 and 3).
    \item \textbf{Appendix \ref{appendix:additional_results}}: Additional Results (cf. Section 3).
\end{itemize}

\vspace{12pt}  

\section{CliKT Construction}
\label{appendix:clikt}

Due to the lack of datasets dedicated to evaluating long-tail biomedical knowledge,
we propose CliKT, a new benchmark specifically designed to evaluate LLMs' performance on long-tail biomedical knowledge. Notably, given that PubMed is a widely used biomedical corpus for pre-training LLMs~\cite{wang2023pre}, which contains over 37 million abstracts of biomedical papers~\cite{wei2013pubtator}, we mainly focus on PubMed data to extract long-tail biomedical knowledge.
Specifically, we first extract knowledge triples from SNOMED CT~\cite{donnelly2006snomed} (\S\ref{subsec:extract_biomedical_knowledge_triples}) to obtain a comprehensive set of biomedical concepts and their relationships.
Next, we employ an entity linking pipeline to map these triples back to their corresponding documents in the PubMed~\cite{roberts2001pubmed} corpus (\S\ref{subsec:mapping_knowledge_triples_to_pubmed}), enabling us to identify whether a triple represents long-tail knowledge based its occurrence in the corpus. 
Finally, we generate question-answer (QA) pairs based on the knowledge triples to evaluate the ability of LLMs to capture the factual knowledge, and conduct a human evaluation to show that our entity linking pipeline accurately identifies relevant documents for the majority of the QA pairs.

\subsection{Extracting Biomedical Knowledge Triples}
\label{subsec:extract_biomedical_knowledge_triples}
We focus on the long-tail biomedical knowledge from the PubMed corpus. 
However, directly extracting such knowledge from the entire corpus is a challenging task~\cite{shetty2021automated,nguyen2021advanced,abdullah2023systematic}. 
Therefore, following previous work~\cite{alghanmi2021probing,fei2021enriching}, we leverage information from existing biomedical knowledge graphs to facilitate more efficient extraction. 
Specifically, we extract all the knowledge triples from SNOMED CT~\cite{donnelly2006snomed}, which is a comprehensive biomedical knowledge graph comprising over 200K triples and widely used for assessing LLMs’ understanding of biomedical knowledge~\cite{meng2021rewire}. Each triple is denoted as (head entity, relation, tail entity), representing the relationship between two entities, e.g., (Type 1 Diabetes, Therapeutic Procedure, Insulin therapy). 

\subsection{Mapping Knowledge Triples to PubMed Documents}
\label{subsec:mapping_knowledge_triples_to_pubmed}

We then develop an entity linking pipeline to map the extracted knowledge triples back to documents in Pubmed~\cite{roberts2001pubmed} to identify long-tail knowledge.
The detailed procedure is as follows: 

\vspace{0.3em} \noindent \textbf{Entity Annotation}. To facilitate the mapping of knowledge triples to specific PubMed documents, we first need to annotate the entities within the PubMed corpus. To this end, we use PubTator~\cite{wei2013pubtator}, a robust web-based text-mining tool that provides automatic annotations of biomedical concepts in PubMed. Following the work of \citet{wei2019pubtator}, we obtain entity annotations within 37 million PubMed abstracts\footnote{The annotated data is available at \url{https://ftp.ncbi.nlm.nih.gov/pub/lu/PubTatorCentral/}}.

\vspace{0.3em} \noindent \textbf{Entity Linking}.
After obtaining annotated entities, the next step is to map the knowledge triples to their corresponding PubMed documents. Previous studies~\cite{elsahar2018t,kandpal2023large} suggest that when the head entity and the tail entity of a knowledge triple co-occur within a document, it is likely that the knowledge represented by the triple is expressed in that document. Based on this observation, we define documents where both the head and tail entities of a knowledge triple co-occur as its \textit{related documents}, and the count of such documents as the \textit{co-occurrence number}.  

To determine whether both the head and tail entities of a triple co-occur in a document, we use SapBERT~\cite{liu2020self}, an effective biomedical entity linking model, to match these entities to those present in the document. 
For instance, given the triple (Hypertension, causes, heart disease) from SNOMED CT, SapBERT can link ``Hypertension'' to its equivalent term ``high blood pressure'' in PubMed, ensuring an accurate match with related documents. 
We iterate through the entire corpus to calculate the co-occurrence number for each triple. We define triples with a low co-occurrence number as long-tail biomedical knowledge. 

\vspace{0.3em}\noindent \textbf{Question Generation.} 
Finally, we generate QA pairs based on the resulting triples to assess the LLMs' ability to capture these knowledge triples. Following~\citet{meng2022locating}, we manually design templates to generate questions using the head entity and the relation, while considering the tail entity as the answer. 
For example, given a triple (Diabetes, treated\_by, Insulin), the corresponding QA pair would be: \textit{Question: What is Diabetes treated by? Answer: Insulin}.

\begin{figure*}[!t]%
    \centering
    \resizebox{1.0\linewidth}{!}{
    \includegraphics[width=0.4\textwidth]{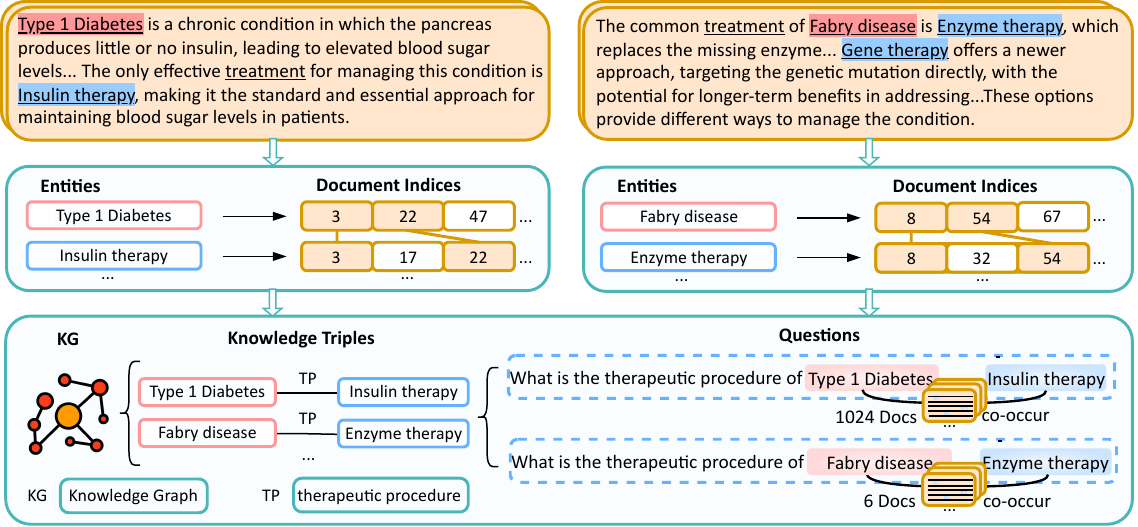}
    }
    \caption{
    The pipeline for identifying long-tail biomedical knowledge consists of a systematic process encompassing document collection, entity linking, knowledge graph traversal, and question generation.
    }
    \label{fig:clikt_construction}
\end{figure*}

\begin{table}[h!]
\centering
\resizebox{0.9\linewidth}{!}{
\begin{tabular}{@{}lp{0.8\linewidth}@{}}
\toprule
\textbf{Relation}                     & \textbf{Template}                                 \\ \midrule
Finding site                          & \textbf{Edit Prompt:} ``The finding site of [SUBJECT] is.'' \\ 
                                      & \textbf{Question:} ``What is the finding site of [SUBJECT]?'' \\
                                      & \textbf{Rephrase:} ``Where is [SUBJECT] typically found?'' \\ \midrule
Associated morphology                 & \textbf{Edit Prompt:} ``The associated morphology of [SUBJECT] is.'' \\
                                      & \textbf{Question:} ``What is the associated morphology of {SUBJECT}?'' \\
                                      & \textbf{Rephrase:} ``Can you describe the morphology associated with [SUBJECT]'' \\ \midrule
Causative agent                       & \textbf{Edit Prompt:} ``The causative agent of [SUBJECT] is'' \\
                                      & \textbf{Question:} ``What is the causative agent of [SUBJECT]?'' \\
                                      & \textbf{Rephrase:} ``Which pathogen causes [SUBJECT]?'' \\ \midrule
Interprets                            & \textbf{Edit Prompt:} ``[SUBJECT] interprets.'' \\
                                      & \textbf{Question:} ``What does [SUBJECT] interprets?'' \\
                                      & \textbf{Rephrase:} ``What is interpreted by [SUBJECT]?'' \\ \midrule
Procedure site                        & \textbf{Edit Prompt:} ``The procedure site of [SUBJECT] is'' \\
                                      & \textbf{Question:} ``What is the indirect procedure site of [SUBJECT]?'' \\
                                      & \textbf{Rephrase:} ``Where is the procedure site for [SUBJECT]?'' \\ \midrule
Pathological process                  & \textbf{Edit Prompt:} ``The pathological process of [SUBJECT] involves.'' \\
                                      & \textbf{Question:} ``What is the pathological process of [SUBJECT]?'' \\
                                      & \textbf{Rephrase:} ``Which pathological process does [SUBJECT] involve?'' \\  \midrule
Due to                                & \textbf{Edit Prompt:} ``[SUBJECT] is due to.'' \\
                                      & \textbf{Question:} ``What is the [SUBJECT] due to?'' \\
                                      & \textbf{Rephrase:} ``What is the cause of [SUBJECT]?'' \\ \midrule
Has active ingredient                 & \textbf{Edit Prompt:} ``The active ingredient of [SUBJECT] is.'' \\
                                      & \textbf{Question:} ``What is the active ingredient of [SUBJECT]?'' \\
                                      & \textbf{Rephrase:} ``What active ingredient does [SUBJECT] have?'' \\ \midrule
Part of                               & \textbf{Edit Prompt:} ``[SUBJECT] is a part of.'' \\
                                      & \textbf{Question:} ``What is the [SUBJECT] a part of?'' \\
                                      & \textbf{Rephrase:} ``To what is [SUBJECT] a part?'' \\ \midrule
Has definitional manifestation        & \textbf{Edit Prompt:} ``The definitional manifestation of [SUBJECT] is.'' \\
                                      & \textbf{Question:} ``What is the definitional manifestation of [SUBJECT]?'' \\
                                      & \textbf{Rephrase:} ``How is [SUBJECT] manifested definitionally?'' \\ \midrule 
Component                             & \textbf{Edit Prompt:} ``The component of [SUBJECT] is.'' \\
                                      & \textbf{Question:} ``What is the component of [SUBJECT]?'' \\
                                      & \textbf{Rephrase:} ``What components does [SUBJECT] consist of?'' \\

\bottomrule
\end{tabular}
}
\caption{Examples of relation templates demonstrate how each relation is transformed into input prompts, which can categorized into three parts: Edit Prompt, Question, and Rephrase. The ``Edit Prompt'' is used for knowledge editing and reliability evaluation, the ``Question'' is designed for knowledge probing, and the ``Rephrase'' is used to assess generalisation metrics. The complete template for all the relations can be found in our github repository. }
\label{table:examples_of_relation_templates}
\end{table}

\section{Experimental Details}
\label{appendix:experimental_details}

\subsection{Details of Large Language Models}
\label{subsec:app_details_of_llms}

We employ two biomedical LLMs and two general-domain LLMs in our experiments:

\begin{itemize} 
\vspace{-0.5em} \item \textbf{BioGPT-Large~\cite{luo2022biogpt}:} A 1.5 billion parameter model from Microsoft, primarily pre-trained on PubMed, excelling in drug discovery and medical record analysis. 
\vspace{-0.5em} \item \textbf{BioMedLM~\cite{bolton2024biomedlm}:} A Stanford-developed model optimised for biomedical tasks, pretrained on PubMed with 2.7 billion parameters, ideal for literature retrieval and information extraction.
\vspace{-0.5em} \item \textbf{Llama2~\cite{touvron2023llama}:} A Meta-developed model with 7 billion parameters, designed for general-purpose language tasks. It has been leveraging large-scale pretraining on diverse datasets, including biomedical corpora. 
\vspace{-0.5em} \item \textbf{GPT-J~\cite{wang2021gpt}:} A 6 billion parameter open-source model by EleutherAI, trained on the Pile dataset, which includes a significant portion of biomedical texts from PubMed. 
\vspace{-0.5em}
\end{itemize}

In addition to the models listed above, we also include results for two recently released models, Llama3~\cite{grattafiori2024llama} and Qwen2.5~\cite{yang2024qwen2}, to provide a broader view of knowledge editing performance across both biomedical-specific and general-purpose LLMs.

\subsection{Details of Evaluation Metrics}
\label{subsec:details_of_evaluation_metrics}

(1) \textbf{Reliability}: This metric measures the average accuracy over a predefined set of input-output pairs $(x_e, y_e)$. It is aimed to evaluate the ability to memorise the form of edit Prompt after knowledge editing.
\begin{equation}
\mathbb{E}_{x'_e, y'_e \sim \{(x_e, y_e)\}} \mathbf{1} \left\{ \operatorname*{argmax}_y f_{\theta_e} (y \mid x'_e) = y'_e \right\}
\label{eq:example}
\end{equation}

(2) \textbf{Generalisation}: Considering that paraphrased sentences are modified accordingly through editing, this metric measures the average accuracy on equivalent neighbours \( R(x_e, y_e) \), where equivalent neighbours are rephrased questions based on the edited knowledge. 

\begin{equation}
\mathbb{E}_{x'_e, y'_e \sim R(x_e, y_e)} \mathbf{1} \left\{ \operatorname*{argmax}_y f_{\theta_e} (y \mid x'_e) = y'_e \right\}
\label{eq:argmax_expectation}
\end{equation}

(3) \textbf{Locality}: 
This metric measures the frequency with which the predictions of the post-edit model remain consistent for out-of-scope neighbors \( O(x_e, y_e) \).
\begin{equation}
\mathbb{E}_{x'_e, y'_e \sim O(x_e, y_e)} \mathbf{1} \left\{ f_{\theta_e} (y \mid x'_e) = f_{\theta} (y \mid x'_e) \right\}
\label{eq:consistency}
\end{equation}




\subsection{Details of Tuning Process}
\label{subsec:details_of_tuning_process}

\section{Additional Results}
\label{appendix:additional_results}

\begin{figure*}[tb]
    \centering
    \subfigure[The performance on BioGPT.]{
        \includegraphics[width=0.48\linewidth]{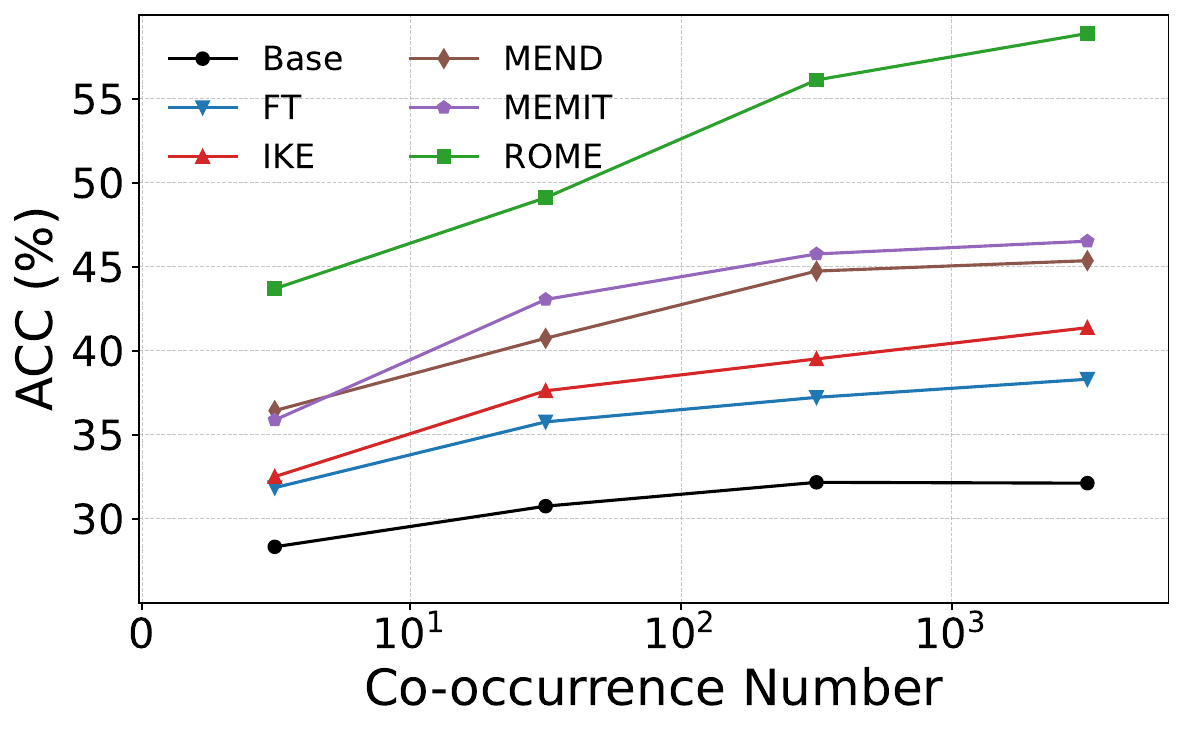}
        \label{fig:biogpt_model_editing_performance}
    }
    \hfill
    \subfigure[The performance on Llama2.]{
        \includegraphics[width=0.48\linewidth]{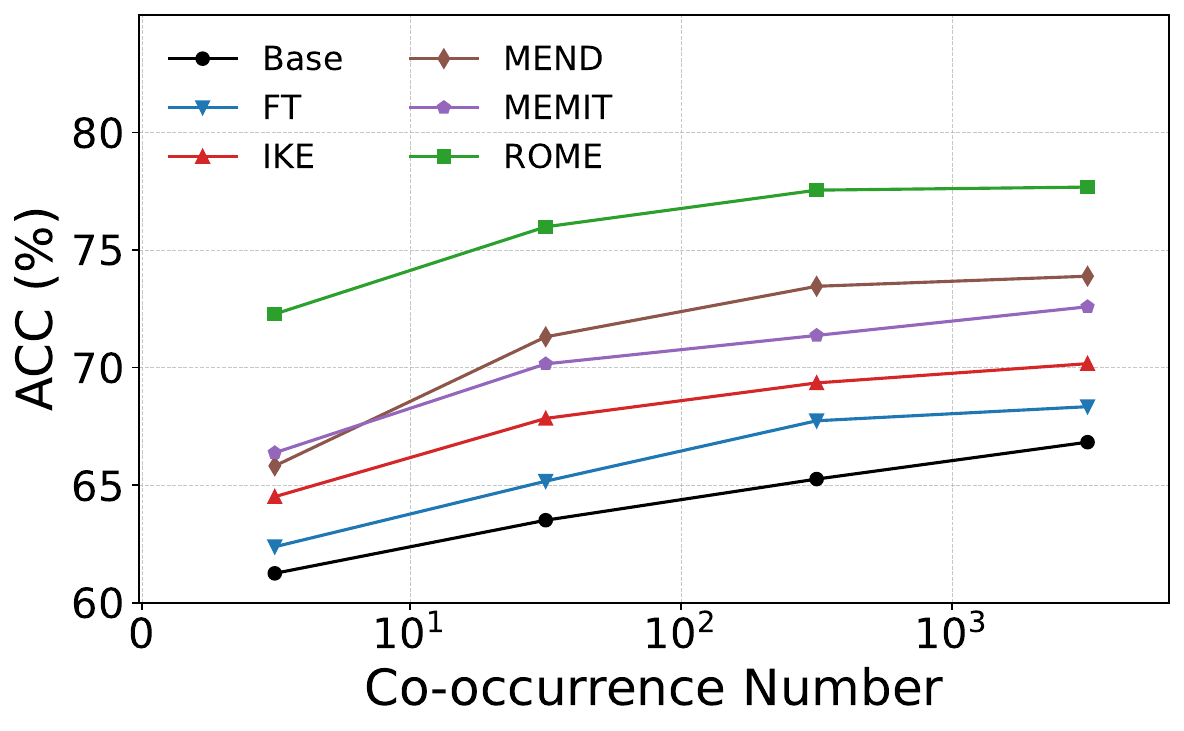}
        \label{fig:llama_model_editing_performance}
    }
    \hfill
    \subfigure[The performance on Llama3.]{
        \includegraphics[width=0.48\linewidth]{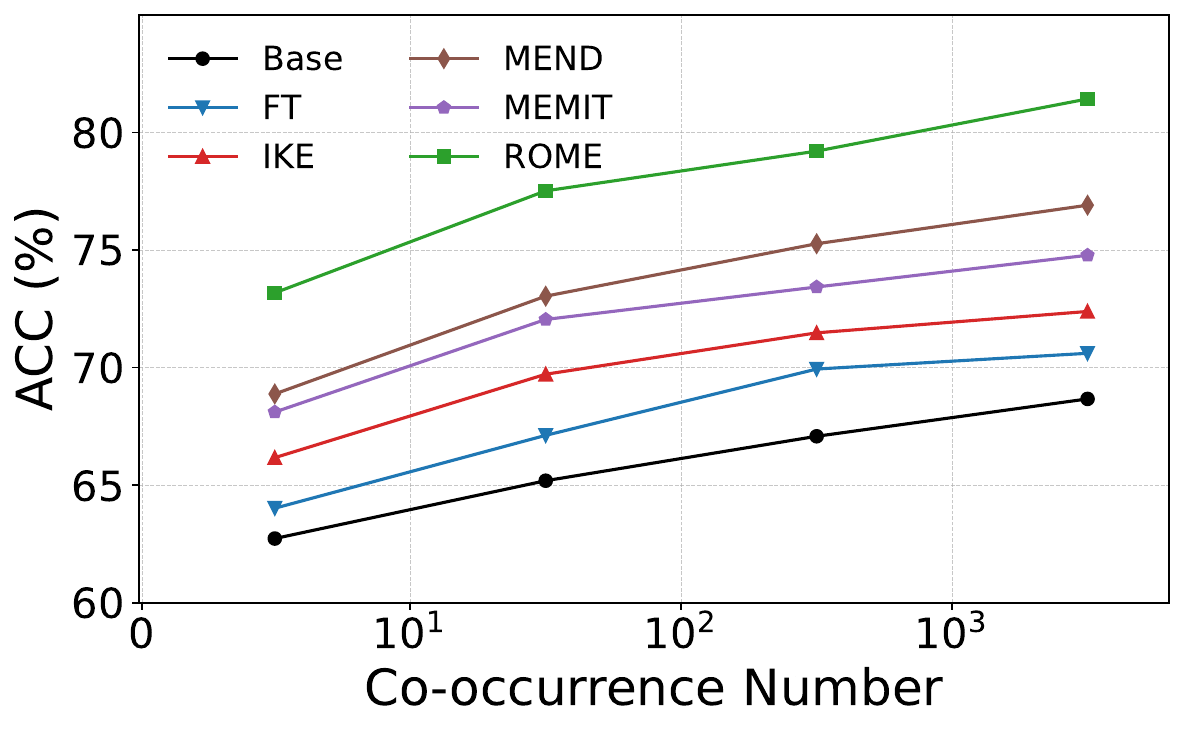}
        \label{fig:llama3_model_editing_performance}
    }
    \hfill
    \subfigure[The performance on Qwen2.5.]{
        \includegraphics[width=0.48\linewidth]{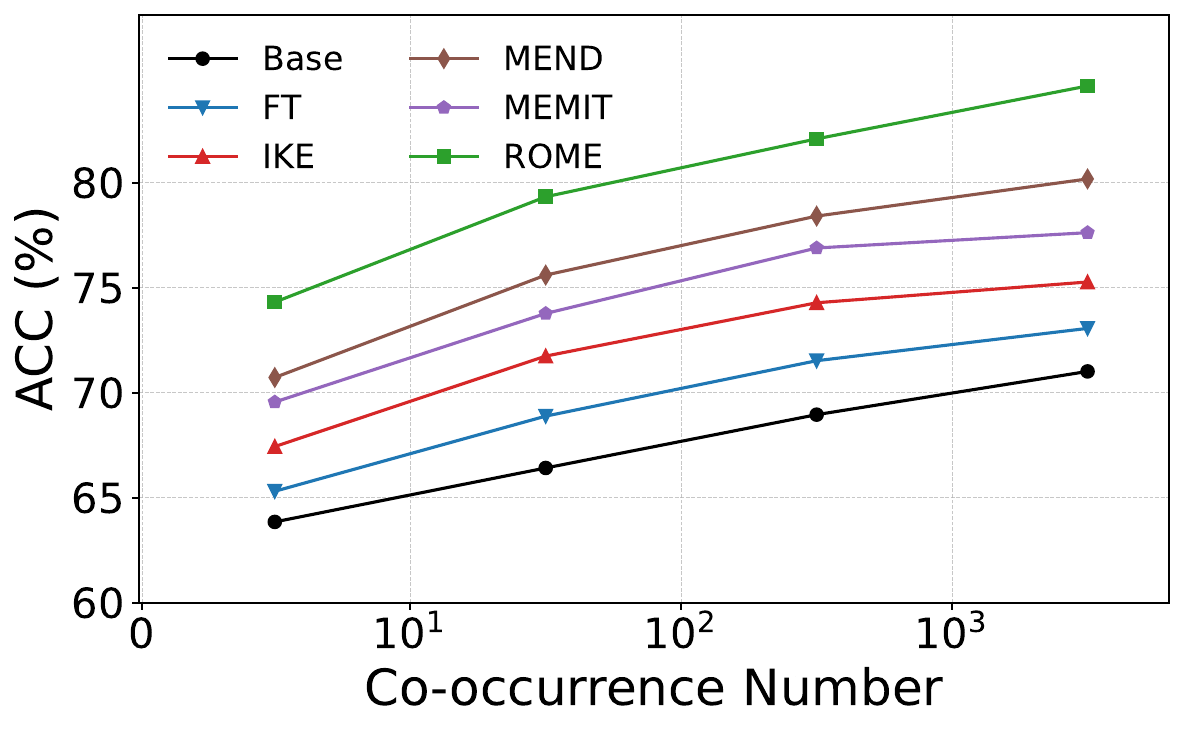}
        \label{fig:qwen2.5_model_editing_performance}
    }
    \caption{The performance of knowledge probing after editing with different editing methods on BioGPT and Llama2, where ``Base'' denotes LLM without editing. }
    \label{fig:additional_editing_performance}
\end{figure*}

We present the performance of knowledge editing on additional base LLMs in this section. In particular, we evaluate the post-edit probing accuracy of BioGPT\cite{luo2022biogpt}, Llama2\cite{touvron2023llama}, Llama3~\cite{grattafiori2024llama}, and Qwen2.5~\cite{yang2024qwen2} using a range of editing methods. The results are shown in Figure~\ref{fig:biogpt_model_editing_performance}, Figure~\ref{fig:llama_model_editing_performance}, Figure~\ref{fig:llama3_model_editing_performance}, and Figure~\ref{fig:qwen2.5_model_editing_performance}, respectively.

\begin{figure*}[tb]%
    \centering
    \resizebox{0.85\linewidth}{!}{
    \includegraphics[width=0.4\textwidth]{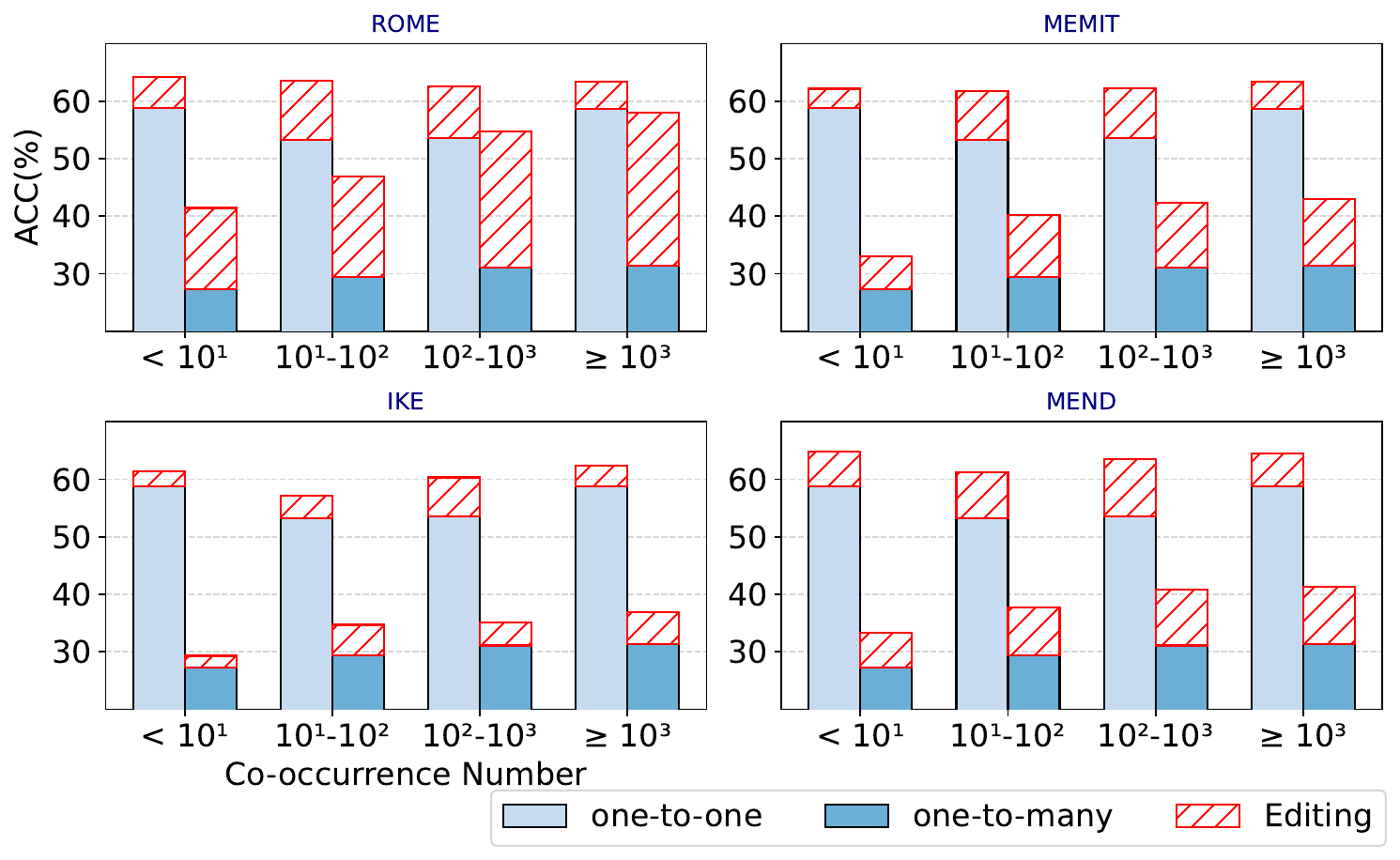}
    }
    \caption{The knowledge probing performance of BioGPT on both one-to-one knowledge and one-to-many knowledge before and after editing. }
    \label{fig:biogpt_dif_relation_editing_performance}
\end{figure*}

\begin{figure*}[tb]%
    \centering
    \resizebox{0.85\linewidth}{!}{
    \includegraphics[width=0.4\textwidth]{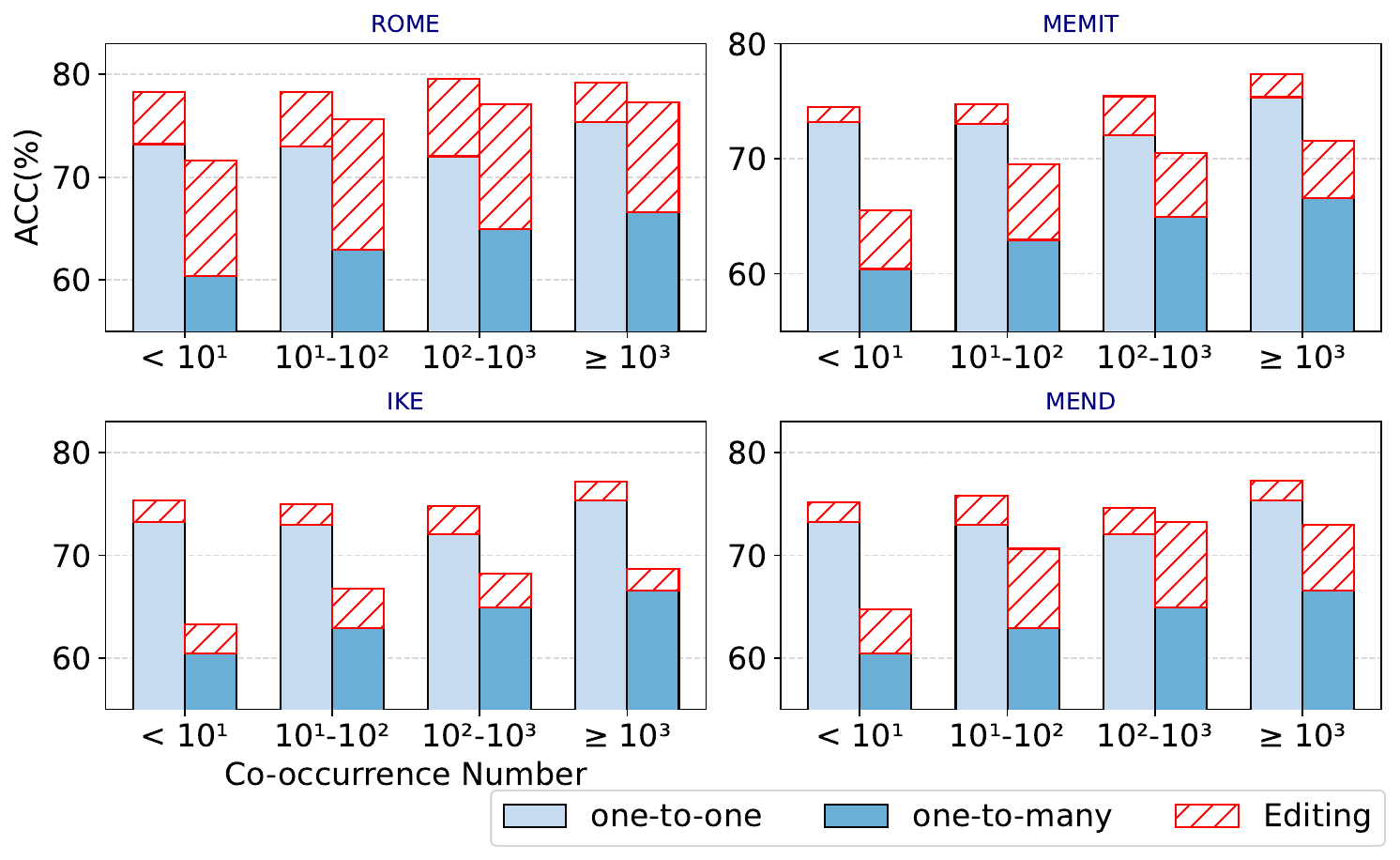}
    }
    \caption{The knowledge probing performance of Llama2 on both one-to-one knowledge and one-to-many knowledge before and after editing. }
    \label{fig:llama_dif_relation_editing_performance}
\end{figure*}

\begin{figure*}[tb]%
    \centering
    \resizebox{0.85\linewidth}{!}{
    \includegraphics[width=0.4\textwidth]{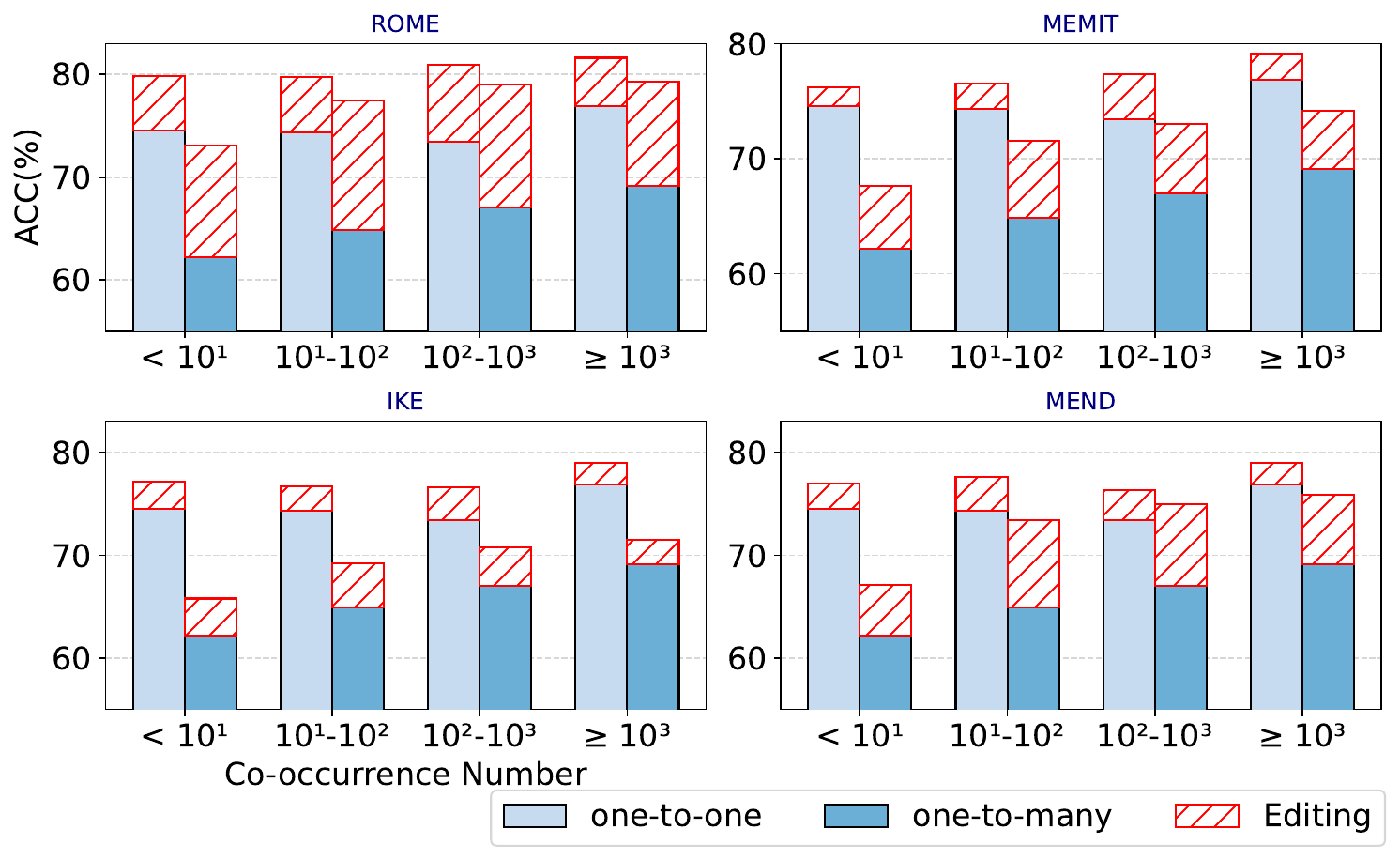}
    }
    \caption{The knowledge probing performance of Llama3 on both one-to-one knowledge and one-to-many knowledge before and after editing. }
    \label{fig:llama3_dif_relation_editing_performance}
\end{figure*}

\begin{figure*}[tb]%
    \centering
    \resizebox{0.85\linewidth}{!}{
    \includegraphics[width=0.4\textwidth]{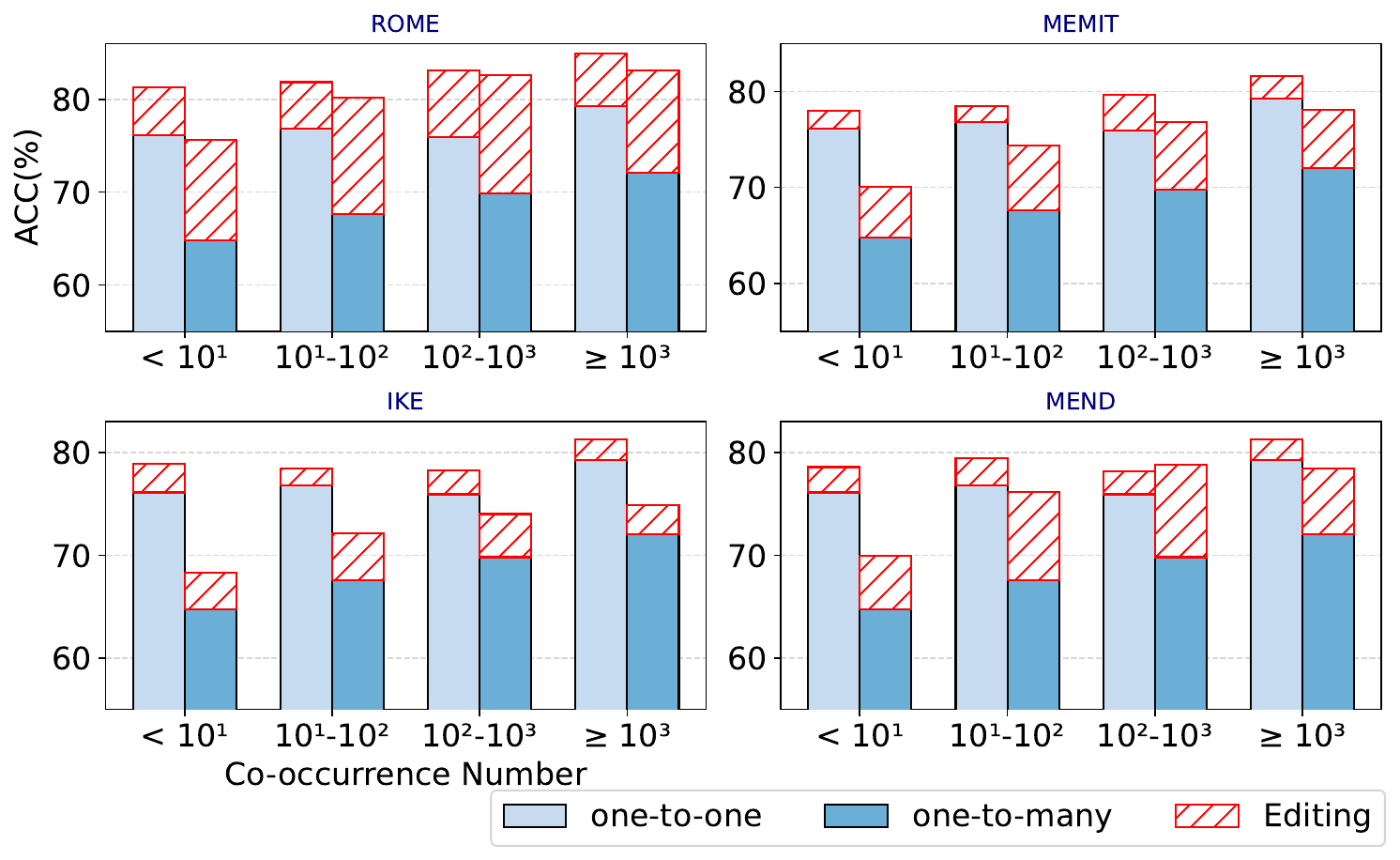}
    }
    \caption{The knowledge probing performance of Llama3 on both one-to-one knowledge and one-to-many knowledge before and after editing. }
    \label{fig:qwen2.5_dif_relation_editing_performance}
\end{figure*}

To further investigate the impact of editing across different types of biomedical knowledge, we also conduct a relation-level analysis for each model. These results are presented in Figure~\ref{fig:biogpt_dif_relation_editing_performance}, Figure~\ref{fig:llama_dif_relation_editing_performance}, Figure~\ref{fig:llama3_dif_relation_editing_performance}, and Figure~\ref{fig:qwen2.5_dif_relation_editing_performance}.

\end{document}